\documentclass{article}

\usepackage{arxiv}
\usepackage{algorithm}
\usepackage{algorithmic}
\usepackage[utf8]{inputenc} 
\usepackage[T1]{fontenc}    
\usepackage{hyperref}       
\usepackage{url}            
\usepackage{booktabs}       
\usepackage{amsfonts}       
\usepackage{nicefrac}       
\usepackage{microtype}      
\usepackage{graphicx}
\usepackage{natbib}
\usepackage{doi}
\usepackage{amsmath,amssymb,amsthm}
\usepackage[utf8]{inputenc}

\usepackage{caption}
\usepackage{subcaption} 
\usepackage{adjustbox}
\usepackage{calc}       

\hypersetup{hidelinks}

\newtheorem{theorem}{Theorem}[section]
\newtheorem{definition}{Definition}[section]
\newtheorem{lemma}{Lemma} 

\title{Low Rank Support Quaternion Matrix Machine}

\author{ \href{https://orcid.org/0009-0007-4962-1130}{\includegraphics[scale=0.06]{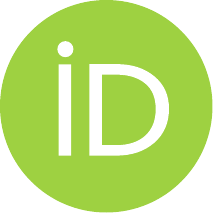}\hspace{1mm}Wang Chen}\\
 	School of Mathematics and Statistics\\
 	Beijing Jiaotong University\\
     Beijing 100044,100044  \\
 	\texttt{24121561@bjtu.edu.cn} \\
	\And
	\href{https://orcid.org/0000-0002-4926-5929}{\includegraphics[scale=0.06]{orcid.pdf}\hspace{1mm}Ziyan Luo}
   \thanks{Corresponding author. This author's work was supported by the National Natural Science Foundation of China (12271022) and National Key Research and Development Program of China (2024YFA1012901).}\\
	 School of Mathematics and Statistics\\
	 Beijing Jiaotong University\\
     Beijing 100044, China  \\
	 \texttt{zyluo@bjtu.edu.cn} \\
     \And
	\href{https://orcid.org/0009-0004-5237-4618}{\includegraphics[scale=0.06]{orcid.pdf}\hspace{1mm}Shuangyue Wang} \\
	 College of Information and Management Science\\
	 Henan Agricultural University\\
     Zhengzhou 450046, China  \\
	 \texttt{sywangmath@163.com} \\
}

\renewcommand{\shorttitle}{LSQMM}

\hypersetup{
pdftitle={ Low Rank Support Quaternion Matrix Machine},
pdfsubject={q-bio.NC, q-bio.QM},
pdfauthor={Wang Chen},
pdfkeywords={First keyword, Second keyword, More},
}

\begin{document}
\maketitle

\begin{abstract}
Input features are conventionally represented as vectors,  matrices, or third order tensors in the real field, for color image classification. 
Inspired by the success of quaternion data modeling for color images in image recovery and denoising tasks, we propose a novel classification method for color image classification, named as the Low-rank Support Quaternion Matrix Machine (LSQMM), in which the RGB channels are treated as pure quaternions to effectively preserve the intrinsic coupling relationships among channels via the quaternion algebra. For the purpose of promoting low-rank structures resulting from strongly correlated color channels, a quaternion nuclear norm regularization term, serving as a natural extension of the conventional matrix nuclear norm to the quaternion domain, is added to the hinge loss in our LSQMM model. An Alternating Direction Method of Multipliers (ADMM)-based iterative algorithm is designed to effectively resolve the proposed quaternion optimization model.  
Experimental results on multiple color image classification datasets demonstrate that our proposed classification approach exhibits advantages in classification accuracy, robustness and computational efficiency, compared to several state-of-the-art methods using support vector machines, support matrix machines, and support tensor machines.
\end{abstract}


\section{Introduction}
Image classification serves as a cornerstone in computer vision and pattern recognition, with far-reaching impacts across diverse domains including medical diagnostics~\cite{esteva2017doctorpic}, autonomous navigation~\cite{chen2017selfdriving}, industrial defect detection~\cite{tabernik2020segmentationdetect}, and remote sensing analysis~\cite{zhu2017deepremotesensoring}. The proliferation of digital imaging technologies has established color images as the dominant data format, offering superior spectral and textural information over grayscale counterparts. Consequently, developing accurate and efficient color image classification methods carries substantial theoretical and practical significance~\cite{plataniotis2000colorimage}.

The Support Vector Machine (SVM)~\cite{cortes1995supportvm}, grounded in statistical learning theory, represents one of the most influential classification paradigms due to its solid theoretical foundation and strong generalization performance. Subsequent enhancements have yielded variants including Fuzzy SVM(FSVM)~\cite{lin2002fuzzysvm}, Twin SVM(TWSVM)~\cite{khemchandani2007twinsvm}, and Fuzzy Twin Bounded Large Margin Distribution Machine(FTBLDM)~\cite{jin2022fuzzyTBLDM}. Nevertheless, these conventional approaches typically require flattening multidimensional image data into one-dimensional vectors. Such vectorization disrupts inherent spatial structures, induces high-dimensionality challenges~\cite{li2006multitrainingdimension}, and disregards critical inter-channel correlations in color imagery.

To address these limitations, researchers have developed models that directly operate on matrix and tensor representations. Inspired by matrix completion theory,~\cite{luo2015SMM} proposed the Support Matrix Machine (SMM), which preserves two-dimensional spatial organization. Subsequent developments include Robust SMM(RSMM)~\cite{zheng2018RSMM}, Optimal Distribution Margin Machine(ODMM)~\cite{yang2024ODMM}, and Sparse SMM(SSMM)~\cite{zheng2018SSMM}, which employ nuclear norm regularization to capture low-rank properties in applications like grayscale classification and EEG analysis. However, SMM-based methods remain inadequate for higher-order color image data. The Support Tensor Machine (STM)~\cite{tao2005supervisedtm} extends this framework to tensor representations, with further contributions including Support Tucker Machine(STuM)~\cite{kotsia2011supporttm} and Support Tensor-Train Machine(TT-MMK)~\cite{kour2023TT-MMK}. Despite these advances, existing tensor methods typically process RGB channels as separate dimensions or through simple concatenation, failing to adequately model complex spectral interactions and nonlinear channel couplings, thus limiting their expressive power for color images.

Quaternion algebra has recently emerged as a powerful alternative for color image representation, offering fundamental advantages over conventional approaches. By encoding RGB channels within the three imaginary components of a pure quaternion matrix (formulated as $R\mathbf{i} + G\mathbf{j} + B\mathbf{k}$), this representation inherently preserves spatial structure while maintaining spectral correlations and phase information through quaternion algebraic operations~\cite{ell2006hypercomplexqutercolor,pei1999quaternionimage}. This establishes a more natural mathematical framework for color image processing. Specifically, in image denoising, quaternion-based filtering methods simultaneously consider the interactions among all color channels, effectively suppressing noise while better preserving edge details and color consistency\cite{yu2019quaterniondenoisy}. In image inpainting tasks, low-rank quaternion matrix completion techniques fully utilize the strong correlations between channels to achieve more accurate and natural reconstruction of missing regions\cite{chen2022colorimageinpainting,jia2019robustimageinpaintign}. Additionally, in fields such as image compression\cite{jin2010quaternionfilitering}, digital watermarking\cite{chen2018quaternionwatermark}, and super-resolution reconstruction\cite{km2022qsrnetsuperrecover}, quaternion methods consistently demonstrate performance superior to traditional methods.

Theoretical and empirical studies indicate that the quaternion algebra can naturally capture phase relations and inter-channel dependencies in color images, which helps to preserve a coherent color structure \cite{shi2007quaternion}. These results support the practical advantages of quaternion-based representations in color image processing. Nevertheless, the systematic integration of quaternion matrix representations into support vector machine frameworks for color image classification has not been explored sufficiently.

By introducing the quaternion matrix representation into the support vector machine learning framework, we propose an innovative Low-rank Support Quaternion Matrix Machine (LSQMM) model that provides a brand new approach for color image classification. To the best of our knowledge, this is the first work that integrates quaternion matrix algebra into the support vector machine framework. The main contributions of this paper are reflected in the following three aspects:

\emph{Theoretical Model Innovation}: A low-rank support quaternion matrix machine model, termed LSQMM, is proposed for color image classification. 
Compared with traditional methods, LSQMM not only maintains the spatial structure of images but, more importantly, fully explores the intrinsic coupling relationships among RGB channels through quaternion algebraic operations.

\emph{Efficient Algorithm Design}: 
An efficient numerical algorithm based on the Alternating Direction Method of Multipliers is designed, with convergence and complexity analysis. 

\emph{Numerical Experimental Verification}: 
Numerical experiments are conducted on six public color image datasets, covering multiple important application scenarios including medical imaging and natural objects. Compared with classical LIBSVM, Support Matrix Machine, and Support Tensor Machine, LSQMM outperforms in both classification accuracy and F1-score. 

The paper is organized as follows: Section~\ref{Preliminaries} introduces basic concepts in quaternion algebra and matrix representations; Section~\ref{sec:3} presents the LSQMM model and its ADMM-based optimization algorithm; Section~\ref{sec:4} conducts numerical experiments to demonstrate the effectiveness of our approach; Section~\ref{sec:5} draws some concluding remarks.

\section{Preliminaries}
\label{Preliminaries}

This section recalls several basic concepts and tools in quaternion matrix theory.

An $m \times n$ quaternion matrix comprises one real and three imaginary components, expressed as:
\begin{equation}
\mathbf{A} = A_0 + A_1 \mathbf{i} + A_2 \mathbf{j} + A_3 \mathbf{k},
\end{equation}
where $A_0, A_1, A_2, A_3 \in \mathbb{R}^{m \times n}$, and the fundamental quaternion units $\mathbf{i},\mathbf{j},\mathbf{k}$ obey the following relations:

$$
\mathbf{i}^2 =\mathbf{j}^2 =\mathbf{k}^2 = -1, \quad \mathbf{i}\mathbf{j} = -\mathbf{j}\mathbf{i} = \mathbf{k}, \quad \mathbf{j}\mathbf{k} = -\mathbf{k}\mathbf{j} = \mathbf{i}, \quad \mathbf{k}\mathbf{i} = -\mathbf{i}\mathbf{k} = \mathbf{j}.
$$

Matrices with zero real component ($A_0 = 0$) are termed pure quaternion matrices. The set of $m \times n$ quaternion matrices is denoted by $\mathbb{Q}^{m \times n}$. For matrices $\mathbf{A}, \mathbf{B} \in \mathbb{Q}^{m \times n}$ and $\mathbf{C} \in \mathbb{Q}^{n \times \ell}$, the matrix addition follows:
\begin{equation}
\mathbf{A}+\mathbf{B} = (A_0 + B_0) + (A_1 + B_1)\mathbf{i} + (A_2 + B_2)\mathbf{j}  + (A_3 + B_3)\mathbf{k},
\end{equation}
while matrix multiplication is defined as:

\begin{equation}
\begin{aligned}
\mathbf{AC} = & (A_0C_0 - A_1C_1 - A_2C_2 - A_3C_3) +
       (A_0C_1 + A_1C_0 + A_2C_3 - A_3C_2)\mathbf{i} + \\
      & (A_0C_2 - A_1C_3 + A_2C_0 + A_3C_1)\mathbf{j} +
      (A_0C_3 + A_1C_2 - A_2C_1 + A_3C_0)\mathbf{k}.
\end{aligned}
\end{equation}

Quaternion conjugation for $\mathbf{a} = a_0 + a_1\mathbf{i} + a_2\mathbf{j} + a_3\mathbf{k}$ is given by $\mathbf{a}^* = a_0 - a_1\mathbf{i} - a_2\mathbf{j} - a_3\mathbf{k}$, with modulus $|\mathbf{a}| = \sqrt{a_0^2 + a_1^2 + a_2^2 + a_3^2}$. Note that quaternion matrix operations follow classical matrix operations, with quaternion arithmetic applied to element-wise multiplications. The identity quaternion matrix matches its real-valued counterpart. A square quaternion matrix $\mathbf{A}$ is unitary if $\mathbf{A^*A} = \mathbf{AA^*} = I$, where $\mathbf{A}^*$ denotes the conjugate transpose. Norm definitions for quaternion vectors and matrices parallel those for real and complex cases.

\begin{definition}[\cite{zhang1997qsvd}]
The maximum number of right linearly independent columns of a quaternion matrix $\mathbf{A} \in \mathbb{Q}^{m \times n}$ is called the rank of $\mathbf{A} $.
\end{definition}

It is clear that the rank of $\mathbf{A} $ can refer to the maximum number of left linearly independent rows of $\mathbf{A} $. Moreover, for any two invertible quaternion matrices $\mathbf{P} $ and $\mathbf{Q} $ of appropriate sizes, $\mathbf{A}$ and $\mathbf{PAQ}$ have the same rank.

\begin{definition}
The inner product of quaternion matrices $\mathbf{X}$ and $\mathbf{Y}$ is defined as:
\[
\langle\mathbf{X},\mathbf{Y}\rangle = \text{Tr}(\mathbf{X}^* \mathbf{Y})
\]
where $\text{Tr}(\cdot)$ denotes the matrix trace operation.
\end{definition}

\begin{theorem}[\textbf{Quaternion Singular Value Decomposition (QSVD)}\cite{zhang1997qsvd}]
Let $\mathbf{A}\in \mathbb{Q}^{m \times n}$ be of rank $r$, there exist unitary quaternion matrices $\mathbf{U} = [u_1, u_2, \dots, u_m] \in \mathbb{Q}^{m \times m}$ and $\mathbf{V} = [v_1, v_2, \dots, v_n] \in \mathbb{Q}^{n \times n}$ satisfying

\begin{equation}
\mathbf{A} = \mathbf{U} \Sigma \mathbf{V}^*,
\end{equation}
where $\Sigma = \operatorname{diag}(\sigma_1, \dots, \sigma_r, 0, \dots, 0) \in \mathbb{R}^{m \times n}$, and $\sigma_i$ ($i = 1, \dots, r$) represent the positive singular values of $\mathbf{A}$.
\end{theorem}

\begin{definition}[\cite{wei2018quaternion}]
Quaternion vector and matrix norms are defined as follows:
\begin{enumerate}
    \item[(i)] Let $\mathbf{x} = [x_i] \in \mathbb{Q}^n$ be a quaternion vector.
    $\|\mathbf{x}\|_1 := \sum_{i=1}^n |x_i|$,
    $\|\mathbf{x}\|_2 := \sqrt{\sum_{i=1}^n |x_i|^2}$, and
    $\|\mathbf{x}\|_\infty := \max_{1 \leq i \leq n} |x_i|$.

    \item[(ii)] Let $\mathbf{A} = [a_{ij}] \in \mathbb{Q}^{n_1 \times n_2}$ be a quaternion matrix.
    The $\mathcal{L}_1$-norm $\|\mathbf{A}\|_1 := \sum_{i=1}^{n_1} \sum_{j=1}^{n_2} |a_{ij}|$,
    the $\infty$-norm $\|\mathbf{A}\|_\infty := \max_{i,j} |a_{ij}|$,
    the Frobenius-norm $\|\mathbf{A}\|_F = \sqrt{\sum_{i=1}^{n_1} \sum_{j=1}^{n_2} |a_{ij}|^2} := \sqrt{\text{Tr}(\mathbf{A^* A})}$,
    the spectral norm $\|\mathbf{A}\| := \max\{\sigma_1, \ldots, \sigma_r\}$, and
    the nuclear norm $\|\mathbf{A}\|_{Q,*} := \sum_{i=1}^r \sigma_i$,
    where $\sigma_1, \ldots, \sigma_r$ are positive singular values of $\mathbf{A}$.

\end{enumerate}
\end{definition}

\section{Low-rank Support Quaternion Matrix Machine}
In this section, we propose the LSQMM model for color image classification and design an efficient ADMM method with the complexity and convergence analysis.
\label{sec:3}

\subsection{The LSQMM Model}
\label{3.1}
For a binary classification problem, given a training dataset $\{\mathbf{X}_i, y_i\}_{i=1}^N$, where $\mathbf{X}_i\in \mathbb{Q}^{m \times n}$ represents the sample quaternion matrix and $y_i \in \{-1,1\}$ denotes the class label, the decision function is defined as:
$f(\mathbf{X}) = \mathrm{Re}\big(\langle \mathbf{W}, \mathbf{X}\rangle + b\big),$
where $\mathbf{W}\in \mathbb{Q}^{m \times n}$ is the weight quaternion matrix and $b\in \mathbb{R}$ is the bias term. Considering the presence of potential outliers, we propose a novel Low-rank Support Quaternion Matrix Machine (LSQMM) model:
\begin{equation}
\begin{aligned}
&\min_{\mathbf{W}\in\mathbb{Q}^{m\times n},\; b\in\mathbb{R},\; \{\xi_i\}}
\quad \frac{1}{2} \|\mathbf{W}\|_F^2 +\lambda\,\|\mathbf{W}\|_{Q,*}\;+\; C\sum_{i=1}^N \xi_i\\[6pt]
&\qquad\quad\text{s.t.}\qquad \qquad y_i\Big(\mathrm{Re}\langle \mathbf{W}, \mathbf{X}_i\rangle + b\Big) \;\ge\; 1 - \xi_i,\quad i=1,\dots,N,\\[4pt]
&\qquad\qquad\qquad\qquad \xi_i \ge 0,\quad i=1,\dots,N,
\end{aligned}
\label{Model-1(1)}
\end{equation}
where \(C>0\) and \(\lambda>0\) are penalty parameters that control the weight of classification violation and the low-rankness in the quaternion matrix to be determined, respectively.

\subsection{Optimization Algorithm}
\label{3.2}
Inspired by the Alternating Direction Method of Multipliers (ADMM) \citep{ADMM} proposed in the real field, we design an efficient quaternion ADMM for solving the above quaternion matrix optimization problem \eqref{Model-1(1)} based on the following reformulation by introducing an auxiliary quaternion matrix variable $\mathbf{Z}\in\mathbb Q^{m\times n}$:  
\begin{equation}
\begin{aligned}
&\min_{\mathbf{W}\in\mathbb Q^{m\times n},\; b\in\mathbb R}
\quad f(\mathbf{W}, b, \mathbf{Z}):= \frac{1}{2} \|\mathbf{W}\|_F^2 +\lambda\,\|\mathbf{Z}\|_{Q,*} \;+\; C\sum_{i=1}^N \ h\left(1-y_i\left(\mathrm{Re}\langle \mathbf{W}, \mathbf{X}_i\rangle + b\right)\right)\\[6pt]
&\qquad\text{s.t.}\qquad\qquad \mathbf{W}=\mathbf{Z},
\end{aligned}
\label{Model-1(3)}
\end{equation}
where $h$ in the hinge loss term is defined by $h(t)=\max\{0,t\}$. Note that the augmented Lagrangian function with respect to problem \eqref{Model-1(3)} takes the form of:
\begin{equation}
L_\rho(\mathbf{W}, b, \mathbf{Z}, \mathbf{U}) = f(\mathbf{W}, b, \mathbf{Z}) - \mathrm{Re} \langle \mathbf{U}, \mathbf{W} - \mathbf{Z} \rangle + \frac{\rho}{2} \| \mathbf{W} - \mathbf{Z}\|_F^2,
\label{Augment-Lagrange}
\end{equation}
where $\mathbf{U}\in\mathbb Q^{m\times n} $ is the Lagrange multiplier, and $ \rho $ is the penalty parameter.
The iteration scheme of the ADMM algorithm is given by 

\begin{equation}
\left\{
\begin{aligned}
&(\mathbf{W}^{k+1},b^{k+1}) = \arg \min_{\mathbf{W},b} L_\rho(\mathbf{W}, b, \mathbf{Z}^k, \mathbf{U}^k), \\
&\mathbf{Z}^{k+1} = \arg \min_\mathbf{{Z} }L_\rho(\mathbf{W}^{k+1}, b^{k+1}, \mathbf{Z}, \mathbf{U}^k), \\
&\mathbf{U}^{k+1} = \mathbf{U}^k - \tau \rho(\mathbf{W}^{k+1} - \mathbf{Z}^{k+1}).
\end{aligned}
\right.
\label{ADMM-Frame}
\end{equation}

\noindent {\bf Update for ($\mathbf{W},b$):} The corresponding subproblem can be cast to as

\begin{equation}
\begin{aligned}
&\min_{\mathbf{W}\in\mathbb Q^{m\times n},\; b\in\mathbb R, \xi\in \mathbb R^N\; }
\quad \frac{1}{2} \|\mathbf{W}\|_F^2 +\; C\sum_{i=1}^N \xi_i- \mathrm{Re} \langle \mathbf{U}^k, \mathbf{W} \rangle + \frac{\rho}{2} \| \mathbf{W} - \mathbf{Z}^k\|_F^2\\
&\qquad\quad\text{s.t.}\qquad \qquad y_i\Big(\mathrm{Re}\langle \mathbf{W}, \mathbf{X}_i\rangle + b\Big) \;\ge\; 1 - \xi_i,\quad i=1,\dots,N,\\[4pt]
&\qquad\qquad\qquad\qquad \xi_i \ge 0,\quad i=1,\dots,N.
\end{aligned}
\label{W-constraint question}
\end{equation}

Its Lagrangian function is:

\begin{equation}
\begin{aligned}
L(\mathbf{W}, b,\xi, \alpha, \gamma) &
=  \frac{1}{2} \| \mathbf{W} \|_F^2  + C \sum_{i=1}^N \xi_i
- \mathrm{Re} \langle \mathbf{U}, \mathbf{W} \rangle + \frac{\rho}{2} \| \mathbf{W} - \mathbf{Z}\|_F^2\\
&- \sum_{i=1}^N \alpha_i\Big(y_i\Big(\mathrm{Re}\langle \mathbf{W}, \mathbf{X}_i\rangle + b\Big)\ -1+\xi_i \Big)- \sum_{i=1}^N \gamma_i\xi_i,
\end{aligned}
\label{W-constraint question Lagrange}
\end{equation}
where nonnegative $\alpha$, $\gamma\in\mathbb R^N$ are the Lagrange multipliers. The KKT conditions can be written as:

\begin{equation}
\left\{
\begin{aligned}
&\nabla_\mathbf{W}L(\mathbf{W},b,\xi,\alpha,\gamma) = 0 \\
&\nabla_b L(\mathbf{W},b,\xi,\alpha,\gamma) = 0 \\
&\nabla_\xi L(\mathbf{W},b,\xi,\alpha,\gamma) = 0 \\
&\alpha_i(y_i(\mathrm{Re} \langle \mathbf{W},\mathbf{X}_i\rangle + b) - 1 + \xi_i) = 0 \\
&\gamma_i \xi_i = 0 \\
&y_i(\mathrm{Re} \langle \mathbf{W},\mathbf{X}_i\rangle + b) \geq 1 - \xi_i \\
&\xi_i \geq 0 , \quad i=1,\dots,N\\
&\alpha_i, \gamma_i \geq 0 ,\quad i=1,\dots,N.
\end{aligned}
\right.
\label{W-constraint question KKT}
\end{equation}

Through the  KKT conditions, we can write the dual problem of the subproblem \eqref{W-constraint question} as shown in :
\begin{equation}
\begin{aligned}
&\max_{\alpha \in \mathbb{R}^N}
\quad \sum_{i=1}^N \left(1-\frac{y_{i} \mathrm{Re} \langle \mathbf{U}^k+\rho\mathbf{Z}^k,\mathbf{{X}}_{i}\rangle}{\rho+1}\right)\alpha_{i} - \frac{1}{2(1+\rho)} \left\| \sum_{i=1}^N \alpha_i y_i \mathbf{X}_i \right\|_F^2 \\
& \text{~~s.t.}  \qquad \sum_{i=1}^N \alpha_i y_i = 0, \\[4pt]
& \qquad \qquad 0 \leq \alpha_i \leq C, \quad i=1,\dots, N,
\end{aligned}
\label{W-constraint question dual}
\end{equation}
%
which is a convex quadratic programming problem. By solving problem \eqref{W-constraint question dual} to achieve an optimal solution $\alpha^*$, along with the KKT conditions \eqref{W-constraint question KKT}, we obtain that 

\begin{equation}
\begin{aligned}
\mathbf{W}^{k+1} = \frac{1}{1+\rho}\Big(\rho \mathbf{Z}^k+\mathbf{U}^k+\sum_{i=1}^N \alpha_i^{*}\ y_i\mathbf{X}_i \Big),\\
b^{k+1} = \frac{1}{|S^{*}|} \sum_{i \in S^{*}} (y_i - \mathrm{Re} \langle \mathbf{W}^{k+1}, \mathbf{X}_i \rangle),
\end{aligned}
\label{W,b optimal}
\end{equation}
where $S^{*} = \{i\in \{1,\ldots, N\}: 0 < \alpha_i^{*}\ < C\}$ is the index set of support matrices determined by the dual problem and $|S^*|$ is the cardinality of $S^*$. As can be seen from the expression of $b^{k+1}$ in \eqref{W,b optimal}, the linear combination of all support quaternion matrices are used, instead of relying on a single support matrix, which will mitigate bias by noises and thereby enhance robustness. 

\noindent {\bf Update for $\mathbf{Z}$:} 
By virtue of the proximal operator of the nuclear norm of quaternion matrices, $\mathbf{Z}^{k+1}$ admits a closed form solution as follows:

\begin{equation}
\begin{aligned}
\mathbf{Z}^{k+1} &= \arg \min_\mathbf{Z} \lambda \| \mathbf{Z} \|_{Q,*} + \mathrm{Re} \langle \mathbf{U}^k, \mathbf{Z} \rangle + \frac{\rho}{2} \| \mathbf{W}^{k+1} - \mathbf{Z} \|_F^2 \\
&= \arg \min_\mathbf{Z} \lambda \| \mathbf{Z} \|_{Q,*} + \frac{\rho}{2} \left\| \mathbf{Z} - \frac{\rho \mathbf{W}^{k+1} - \mathbf{U}^k}{\rho} \right\|_F^2 \\
&= \text{Prox}_{\frac{\lambda}{\rho} \| \cdot \|_{Q,*}} \left( \mathbf{W}^{k+1}-\frac{ \mathbf{U}^k}{\rho} \right)
\end{aligned}
\label{Z-solution}
\end{equation}

\par Therefore, the whole scheme for solving LSQMM in \eqref{Model-1(1)} is summarized in Algorithm \ref{ADMM Algorithm}. 

\begin{algorithm}[H]
\caption{Quaternion ADMM for LSQMM}
\textbf{Input:} Samples \( \{\mathbf{X}_i,y_i\}_{i=1}^N \), parameters \( C, \lambda, \rho>0, \tau\in (0, 1.618]\). Set $k=0$. \\
\textbf{Initialize:} \( \mathbf{W}^0 = \mathbf{0} \), \( \mathbf{Z}^0 = \mathbf{0} \), \( \mathbf{U}^0 = \mathbf{0} \), \( b^0 = 0 \). \\
\textbf{While not converged do} \\
1. Update \( \mathbf{W}^{k+1},b^{k+1} \) according to \eqref{W,b optimal}: \\

2. Update \( \mathbf{Z}^{k+1} \) according to \eqref{Z-solution}: \\

3. Update \( \mathbf{U}^{k+1} = \mathbf{U}^k - \tau\rho (\mathbf{W}^{k+1} - \mathbf{Z}^{k+1}) \): \\

\textbf{End while}
\label{ADMM Algorithm}
\end{algorithm}

\subsection{Complexity Analysis}

Let \(N\) be the number of samples, and \(m, n\) be the dimensions of the input images. In the small-sample high-dimensional image classification scenario (\(N \ll m n\)), the ADMM solution process of SQMM primarily consists of preprocessing, dual QP solution ($\mathbf W$-subproblem), $b$-subproblem, $\mathbf Z$-subproblem, and $\mathbf U$-update. The preprocessing requires constructing the \(N \times N\) Gram matrix \(K\), with complexity \(O(N^2 m n)\). In each ADMM iteration, the $\mathbf W$-subproblem is solved using the OSQP package\footnote{https://github.com/osqp/osqp} to address the dual convex quadratic programming problem  \eqref{W-constraint question dual}, with internal iterations \(I\) and total complexity \(O(N^3 + I N^2 + N m n)\); the $b$-subproblem is of complexity \(O(N m n)\); the $\mathbf Z$-subproblem relies on the quaternion nuclear norm proximal operator, requiring quaternion SVD computation, which is equivalent to singular value decomposition of a \(4m \times 4n\) real matrix with complexity \(O(m n \max(m, n))\). The ${\mathbf U}$-update involves matrix operations with complexity \(O(m n)\).
When \(N \ll m n\), we have \(N^3 \ll N m n \ll m n \max(m, n)\). With the preprocessing complexity and \(T\) iterations, the overall time complexity is \(O\left(N^{2} m n + T \cdot m n \max(m, n)\right)\).

\begin{table}[h]
\centering
\caption{Time Complexity Comparison of Different Algorithms }
\label{tab:complexity_comparison}
\begin{tabular}{lccc}
\toprule
\textbf{Algorithm} & \textbf{Reference} & \textbf{Data Type} & \textbf{Time Complexity} \\
\midrule
LIBSVM & \cite{chang2011Libsvm} & $\mathbf{x}\in \mathbb{R}^{3mn}$ & $O(N^2mn)$ \\
SMM & \cite{luo2015SMM} & $\mathbf{X}\in \mathbb{R}^{m \times 3n}$ & $O(N^2mn + T \cdot mn\max(m,n))$ \\
TT-MMK & \cite{kour2023TT-MMK} & $\mathcal{X}\in \mathbb{R}^{m \times n \times 3}$ & $O(Nmnr + N^2r^3(m + n))$ \\
MLRSTM & \cite{yang2025LSTM} & $\mathcal{X}\in \mathbb{R}^{m \times n \times 3}$ & $O(T \cdot (N^2mn + mn\min(m,n)))$ \\
LSQMM & Ours & $\mathbf{X}\in \mathbb{Q}^{m \times n}$ & $O(N^2mn + T \cdot mn\max(m,n))$ \\
\bottomrule
\end{tabular}
\end{table}

Table~\ref{tab:complexity_comparison} presents a theoretical comparison of time complexity among different algorithms .The analysis reveals that LIBSVM, while efficient with complexity $O(N^2mn)$, vectorizes the input data and thus fails to preserve the inherent spatial structure of images. Tensor-based methods (TT-MMK and MLRSTM) explicitly model the third-order tensor structure but introduce additional computational overhead related to tensor decomposition ranks. Both SMM and our proposed LSQMM maintain matrix representations while effectively handling color information, with LSQMM further leveraging quaternion algebra to preserve color channel correlations. 

\subsection{Convergence Analysis}

This section aims to analyze the convergence of the ADMM algorithm proposed in Section \ref{sec:3}. Traditional ADMM convergence is  primarily established in the real field. We will obtain the convergence of Algorithm \ref{ADMM Algorithm} by 
transforming LSQMM into a standard convex optimization problem in the real field through a norm-preserving real isomorphism mapping. This equivalent real problem satisfies the convergence conditions of the ADMM algorithm, thereby guaranteeing the convergence of our proposed quaternion ADMM algorithm.

\begin{lemma}[Real Representation of Quaternion Matrices \cite{zhang1997qsvd,wei2018quaternion}]
\label{lem:real_equivalence}
Define a mapping $\Psi: \mathbb{Q}^{m \times n} \to \mathbb{R}^{4m \times 4n}$ 
by 
\[
\Psi(\mathbf{W}) =
\begin{pmatrix}
W_0 & -W_1 & -W_2 & -W_3 \\
W_1 & W_0 & -W_3 & W_2 \\
W_2 & W_3 & W_0 & -W_1 \\
W_3 & -W_2 & W_1 & W_0
\end{pmatrix}
\in  \mathbb{R}^{4m \times 4n}, ~~~\forall ~\mathbf{W}=W_0 + W_1 \mathbf{i} + W_2 \mathbf{j} + W_3 \mathbf{k} \in \mathbb{Q}^{m \times n}.
\]
Then there exist positive real scalars $c_1, c_2, c_3$ such that

\begin{itemize}
    \item[(i)] Frobenius Norm: $||\mathbf{W}||_F^2 = c_1 ||\Psi(\mathbf{W})||_F^2$
    \item[(ii)] Real Inner Product: $\mathrm{Re}(\langle \mathbf{W}, \mathbf{X} \rangle) = c_2 \langle \Psi(\mathbf{W}), \Psi(\mathbf{X}) \rangle_F$, where $\langle \cdot, \cdot \rangle_F$ is the standard real matrix inner product
    \item[(iii)] Nuclear Norm: $||\mathbf{W}||_{Q,*} = c_3 ||\Psi(\mathbf{W})||_{*}$, where $||\cdot||_{*}$ is the standard real matrix nuclear norm
\end{itemize}
\end{lemma}

Utilizing the isomorphism mapping $\Psi$ from Lemma \ref{lem:real_equivalence}, and letting $\mathcal{W} = \Psi(\mathbf{W})$, $\mathcal{Z} = \Psi(\mathbf{Z})$, $\mathcal{X}_i = \Psi(\mathbf{X}_i)$, the LSQMM model is equivalent to the following model defined in the real field:
\begin{equation}\label{real-counterpart}
 \min_{\mathcal{W}, \mathcal{Z}, b} \{ G(\mathcal{W}, b) + H(\mathcal{Z}):  \mathcal{W} = \mathcal{Z}\},
\end{equation}
where the constants have been incorporated into the objective function with
\[
G(\mathcal{W}, b) = \frac{1}{2}||\mathcal{W}||_{F}^{2} + C'\sum_{i=1}^{N}h(1-y_{i}(\langle \mathcal{W},\mathcal{X}_{i}\rangle_F + b))
~~{\text{and}~~}
H(\mathcal{Z}) = \lambda'||\mathcal{Z}||_{*}.
\]
Problem \eqref{real-counterpart} is a standard two-block separable convex optimization problem with a strongly convex objective function $G(\mathcal{W}, b)$ a linear constraint $\mathcal{W} = \mathcal{Z}$. The convergence of the corresponding ADMM has been established (see, e.g.,\cite{ADMM}). Due to the equivalence as established in Lemma \ref{lem:real_equivalence}, one can routinely show the convergence of the quaternion ADMM in Algorithm \ref{ADMM Algorithm}.

\section{Numerical Experiments}
\label{sec:4}
To validate the effectiveness of our proposed approach in binary color image classification, we conduct numerical experiments on real-world color image datasets, by contrast with other four state-of-the art approaches including the classical Support Vector Machine LIBSVM \cite{chang2011Libsvm}, Support Matrix Machine SMM \cite{luo2015SMM} based on real matrix inputs, TT-MMK \cite{luo2015SMM} based on tensor-train decomposition, and the multi-class low rank support tensor machine MLRSTM \cite{yang2025LSTM}.
The choices of parameters for all compared algorithms follow the default settings provided in their original authors' MATLAB packages or papers. Considering the dimensionality of the datasets used, the tensor-train ranks for TT-MMK were chosen from $\{1,2,3\}$, while MLRSTM - originally designed for multi-class tasks - was manually adapted for binary classification. All numerical experiments are conducted on a computer with AMD Ryzen5 4600H, Radeon Graphics @3.00 GHz CPU, using Python 3.10 for programming and computation.

The stopping criterion in Algorithm \ref{ADMM Algorithm} is chosen as 
\[
\frac{\|\mathbf{W}^k - \mathbf{Z}^k\|_F}{\max\{\|\mathbf{W}^k\|_F, \|\mathbf{Z}^k\|_F\}} < 1e-3, {\text{~~or~~}} k = 1000.
\]

The nuclear norm regularization parameter $\lambda$ is selected from the set  $\{10^{-4}, 10^{-3}, \ldots, 1\}$, and the soft-margin parameter $C$ is chosen from $\{10^{-3},10^{-2}, 10^{-1}, \ldots, 10^{2},10^{3}\}$, and the dual stepsize $\tau$ is simply chosen as $1$. All experiments have employed 5-fold cross-validation, with all computations repeated 10 times. Statistical measures (mean and standard deviation) of accuracy and F1-score are recorded.

\subsection{Dataset Description}
A color image with a resolution of $ m \times n$ pixels is represented by an  $ m \times n $ quaternion matrix \( \mathbf{X}\in \mathbb{Q}^{m \times n} \), specifically as follows:

\begin{equation}
\mathbf{X}_{ij} = R_{ij}\mathbf{i} + G_{ij}\mathbf{j} + B_{ij}\mathbf{k}, \quad 1 \leq i \leq m, 1 \leq j \leq n,
\end{equation}
where \( R_{ij} \), \( G_{ij} \), and \( B_{ij} \) are the red, green, and blue pixel values at position \( (i, j) \) in the image. We collected six real-world datasets  to evaluate the proposed LSQMM.

\begin{figure}[!htbp]
  \centering

  \begin{tabular}{@{}ccccc@{}}
    \makebox[0.15\textwidth][c]{%
      \includegraphics[width=\linewidth,height=2.6cm,keepaspectratio,clip]{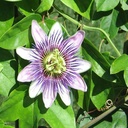}} & 
    \makebox[0.15\textwidth][c]{%
      \includegraphics[width=\linewidth,height=2.6cm,keepaspectratio,clip]{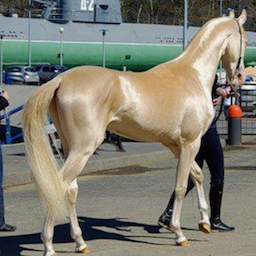}} &
    \makebox[0.15\textwidth][c]{%
      \includegraphics[width=\linewidth,height=2.6cm,keepaspectratio,clip]{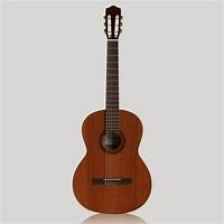}} &
    \makebox[0.15\textwidth][c]{%
      \includegraphics[width=\linewidth,height=2.6cm,keepaspectratio,clip]{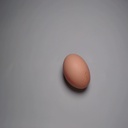}} \\
    
    \makebox[0.15\textwidth][c]{%
      \includegraphics[width=\linewidth,height=2.6cm,keepaspectratio,clip]{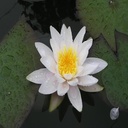}} & 

    \makebox[0.15\textwidth][c]{%
      \includegraphics[width=\linewidth,height=2.6cm,keepaspectratio,clip]{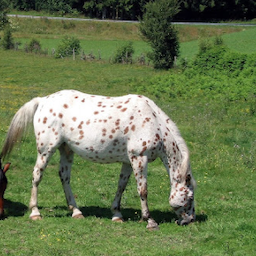}} &
    \makebox[0.15\textwidth][c]{%
      \includegraphics[width=\linewidth,height=2.6cm,keepaspectratio,clip]{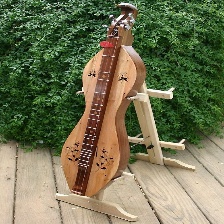}} &
    \makebox[0.15\textwidth][c]{%
      \includegraphics[width=\linewidth,height=2.6cm,keepaspectratio,clip]{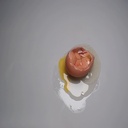}} \\
  \end{tabular}

  \vspace{6pt}
  \caption{\small The image examples  from the proposed dataset.}
  \label{fig:datasets-samples}
\end{figure}

\begin{itemize}
    \item \textbf{Eye Disease Dataset}\footnote{\url{https://www.kaggle.com/datasets/gunavenkatdoddi/eye-diseases-classification}} Contains retinal images from four diagnostic categories. We use 100 images each of Diabetic Retinopathy and Glaucoma, resized to $256\times256$ pixels. Color patterns in blood vessels provide important diagnostic information.

    \item \textbf{Horse Breed Dataset}\footnote{\url{https://www.kaggle.com/datasets/olgabelitskaya/horse-breeds}} Includes seven horse breeds. For binary classification, we selected two visually distinct breeds where coat color and texture are key discriminative features.

    \item \textbf{Musical Instrument Dataset}\footnote{\url{https://www.kaggle.com/datasets/gpiosenka/musical-instruments-image-classification}} Comprises 9,285 images across 10 instrument types. We randomly selected 100 images from two categories and resized them to $112\times112$ pixels. Surface materials and color schemes aid in instrument recognition.

    \item \textbf{Skin Cancer Dataset}\footnote{\url{https://www.kaggle.com/datasets/nodoubttome/skin-cancer9-classesisic}} Provides dermoscopic images from ISIC. Our binary task uses 100 images each of dermatofibroma and seborrheic keratosis ($300\times225$ pixels). Lesion color and texture are critical for diagnosis.

    \item \textbf{Broken Eggs Dataset}\footnote{\url{https://www.kaggle.com/datasets/frankpereny/broken-eggs}} Contains images of eggs in three conditions. We use 100 broken and 100 intact eggs ($128\times128$ pixels). The color contrast between shell and interior provides strong discriminative cues.

    \item \textbf{Oxford Flowers Dataset} Offers 8,189 images of 102 flower species \cite{Oxford}. We selected two categories for binary classification ($128\times128$ pixels). Flower color and petal texture are primary discriminative features.
\end{itemize}

\textbf{Remark on Color Information:} The selected datasets highlight the significance of color features in image classification. In medical domains (Eye Disease, Skin Cancer), color serves as a direct biomarker for pathological conditions. In object recognition tasks (Musical Instruments, Horse Breeds, Flowers, Eggs), color and texture often provide the most salient cues for distinguishing between categories, making these datasets well-suited for evaluating models that effectively leverage chromatic information.

\begin{table}[htbp]
\centering
\caption{Summary of six real-world datasets.}
\label{datasets}
\begin{tabular}{|l|c|c|c|}
\hline
Dataset & Dimension & Size & Category \\ \hline
Eye Disease & $256 \times 256 \times 3$ & 100:100 & 2 \\ \hline
Horse Breed& $256 \times 256 \times 3$ & 100:100 & 2\\ \hline
Music Instrument & $112 \times 112 \times 3$ & 100:100 & 2 \\ \hline
Skin Cancer & $300 \times 225 \times 3$ & 100:100 & 2 \\ \hline
Broken Eggs & $128 \times 128 \times 3$ & 160:200 & 2 \\ \hline
Oxford Flowers & $128 \times 128 \times 3$ & 250:190 & 2 \\ \hline
\end{tabular}

\end{table}

We summarize the main information of these datasets in Table \ref{datasets}. For the above color image datasets, we directly used pixels as input features without any special feature processing.

\subsection{Classification Accuracy}
\label{subsec:experimental_analysis}
Experimental results in terms of accuracy and F1 scores on the aforementioned six datasets are collected in Tables \ref{tab:accuracy_comparison} and \ref{tab:f1_comparison} as below.

\begin{table}[htbp]
\centering
\caption{Accuracy of different methods.}
\label{tab:accuracy_comparison}
\begin{tabular}{lcccccc}
\toprule
Dataset & LIBSVM & SMM & TT-MMK & MLRSTM & LSQMM \\
\midrule
Eye Disease & $86.67\pm0.027$ & $95.00\pm0.022$ & $91.60\pm0.016$ & \textbf{95.50$\pm$0.037} & \textbf{95.50$\pm$0.036} \\
Horse Breed & $77.28\pm0.018$ & $72.25\pm0.030$ & $76.23\pm0.014$ & $80.00\pm0.025$ & \textbf{81.50$\pm$0.053} \\
Music Instrument & $63.70\pm0.025$ & $64.50\pm0.089$ & $77.20\pm0.005$ & $71.00\pm0.038$ & \textbf{80.13$\pm$0.038} \\
Skin Cancer & $90.06\pm0.013$ & $93.00\pm0.032$ & \textbf{95.09$\pm$0.007} & $95.00\pm0.034$ & \textbf{95.09$\pm$0.007} \\
Broken Eggs & $53.17\pm0.024$ & $63.33\pm0.064$ & $75.24\pm0.061$ & $79.17\pm0.041$ & \textbf{80.90$\pm$0.040} \\
Oxford Flowers & $86.51\pm0.012$ & $87.95\pm0.036$ & $86.59\pm0.036$ & \textbf{90.28$\pm$0.031} & \textbf{90.28$\pm$0.016} \\
\bottomrule
\end{tabular}
\end{table}

\begin{table}[htbp]
\centering
\caption{F1-Score of different methods.}
\label{tab:f1_comparison}
\begin{tabular}{lcccccc}
\toprule
Dataset & LIBSVM & SMM & TT-MMK& MLRSTM &  LSQMM \\
\midrule
Eye Disease & $87.00\pm0.034$ &  $94.25\pm0.025$ & $90.60\pm0.009$ & 94.99$\pm$0.018 &\textbf{95.47$\pm$0.034} \\
Horse Breed & $78.50\pm0.031$ & $72.16\pm0.029$ & $76.22\pm0.014$ & 79.96$\pm$0.025 &\textbf{81.10$\pm$0.053}\\
Music Instrument & $71.06\pm0.015$ & $62.37\pm0.088$ & $77.20\pm0.005$ & 70.41$\pm$0.037&\textbf{80.71$\pm$0.034}\\
Skin Cancer & $90.06\pm0.010$ & $93.03\pm0.030$ & \textbf{95.08$\pm$0.008} & 94.98$\pm$0.036&\textbf{95.08$\pm$0.007}\\
Broken Eggs & $0.00\pm0.000$ & $63.05\pm0.043$ & $64.93\pm0.006$ & 76.17$\pm$0.040 &\textbf{76.34$\pm$0.034}\\
Oxford Flowers & $88.83\pm0.015$ & $89.03\pm0.035$ & $85.26\pm0.081$ & 90.28$\pm$0.032&\textbf{91.68$\pm$0.016}\\
\bottomrule
\end{tabular}
\end{table}

As shown in Table~\ref{tab:accuracy_comparison} and Table~\ref{tab:f1_comparison}, LSQMM achieves the highest or tied highest accuracy and F1-score on all six datasets. Particularly on the Music Instrument and Broken Eggs datasets, LSQMM improves accuracy by 16.43\% and 27.73\%, respectively, compared to the traditional Support Vector Machine (LIBSVM), with especially notable F1-score improvements (+9.65\% and +76.34\%). Even on the Oxford Flowers dataset, LSQMM performs comparably to the optimal Support Tensor Machine (MLRSTM), demonstrating strong robustness.

The core advantage of LSQMM stems from the inherent compatibility of quaternion matrix representation:
By treating color images as pure quaternion matrices (zero real part) in \( \mathbb{Q}^{m \times n} \), where the \( i, j, k \) components correspond to R, G, B channels respectively, LSQMM not only completely preserves the spatial topology of images but also naturally models inter-channel dependencies and order sensitivity through the non-commutativity of quaternion multiplication. Combined with nuclear norm regularization, LSQMM further learns low-rank features, effectively suppressing overfitting in small-sample high-dimensional scenarios and enhancing generalization performance.

\subsection{Runtime Efficiency}

This subsection further compares the actual training times of different methods across six color image datasets, with results presented in Table~\ref{tab:time_comparison}. All experiments are conducted under uniform hardware conditions, with reported values representing the average of five independent runs. The training time analysis reveals that TT-MMK incurs significantly higher computational costs compared to other methods, primarily due to the expensive tensor train decomposition operations required for all samples. In contrast, LIBSVM achieves the shortest runtime as it does not utilize structural information for any additional computations. The computational efficiency of our method remains moderate among the compared approaches.

\begin{table}[htbp]
\centering
\caption{Training time (seconds) of different methods.}
\label{tab:time_comparison}
\begin{tabular}{lcccccc}
\toprule
Dataset & LIBSVM & SMM & TT-MMK &  MLRSTM &LSQMM \\
\midrule
Eye Disease & $2.24$ & $46.87$ & $66.69$ & $3.77$ & $6.08$  \\
Horse Breed & $2.41$  & $48.12$ & $61.64$ &$3.61$ & $5.47$  \\
Music Instrument & $1.82$  & $8.35$ & $53.22$ &$3.37$ & $2.54$  \\
Skin Cancer & $2.29$  & $47.50$ & $57.65$ &$3.19$ & $5.38$  \\
Broken Eggs & $5.81$  & $16.62$ & $178.83$ &$3.43$ & $2.65$  \\
Oxford Flowers & $6.10$  & $19.09$ & $291.14$ &$4.54$ & $3.15$  \\
\bottomrule
\end{tabular}
\end{table}

Although LSQMM requires higher computational cost than some baseline methods on certain datasets, the performance improvements justify its effectiveness. For instance, on the Horse Breed dataset, LSQMM's training time (5.47 seconds) is slightly longer than MLRSTM (3.61 seconds), but it achieves 1.50\% higher accuracy and 1.14\% improvement in F1-score. It is noteworthy that on the Oxford Flowers dataset, LSQMM achieves an accuracy of $90.28\pm0.016\%$ with only 3.15 seconds, outperforming MLRSTM which requires 4.54 seconds.

The runtime analysis demonstrates that LSQMM effectively preserves color image structural information through quaternion matrix representation, achieving performance improvements within acceptable computational costs. This provides a balanced solution between performance and efficiency for small-sample high-dimensional color image classification problems.

\subsection{Noise Robustness }

This experiment evaluates the robustness of different methods by introducing Gaussian white noise with varying ratios to the Eye Disease dataset. As shown in Table~\ref{tab:accuracy with noise_comparison}, as the noise ratio $R$ increases from 0.1 to 1.0, all methods exhibit a declining trend in accuracy, with LSQMM demonstrating the best noise robustness. In the table, the best results are bolded and the second-best results are underlined. As before, all experiments are conducted five times to obtain stable results.

\begin{table}[htbp]
\centering
\caption{Accuracy (percentage) of different methods under gaussian white noise ( R is noise ratio).}
\label{tab:accuracy with noise_comparison}
\begin{tabular}{lcccccc}
\toprule
Dataset & & LIBSVM & SMM & TT-MMK & MLRSTM &  LSQMM \\
\midrule
 & R=0.1 & 87.20$\pm$0.015 & \underline{95.00$\pm$0.047} & 89.90$\pm$0.012& 94.50$\pm$0.037 &    \textbf{95.00$\pm$0.012}  \\
& R=0.2 & 84.60$\pm$0.016 & 92.90$\pm$0.021 & 89.10$\pm$0.011 &\underline{93.00$\pm$0.026} &    \textbf{93.00$\pm$0.012}  \\
& R=0.3 & 83.05$\pm$0.018 & \textbf{92.50$\pm$0.038} & 88.90$\pm$0.012 &91.50$\pm$0.026 &    \underline{91.50$\pm$0.011} \\
& R=0.4 & 80.15$\pm$0.018 & 89.50$\pm$0.048 & 87.50$\pm$0.012& \underline{90.00$\pm$0.028} &   \textbf{90.50$\pm$0.010}  \\
Eye Disease& R=0.5 & 78.05$\pm$0.021 & 88.50$\pm$0.065 & 87.60$\pm$0.007&\textbf{90.00$\pm$0.031} & \underline{89.80$\pm$0.011} \\
& R=0.6 & 76.50$\pm$0.015 & 87.00$\pm$0.054 & 86.30$\pm$0.011 &\underline{87.80$\pm$0.037} &    \textbf{87.90$\pm$0.014}  \\
& R=0.7 & 72.50$\pm$0.012 & \underline{86.50$\pm$0.084} & 85.10$\pm$0.019 &85.50$\pm$0.074 &    \textbf{87.00$\pm$0.011}  \\
& R=0.8 & 62.00$\pm$0.014 & 85.00$\pm$0.081 & \underline{85.00$\pm$0.060} &85.00$\pm$0.067&    \textbf{85.00$\pm$0.016}  \\
& R=0.9 & 54.05$\pm$0.030 & \underline{84.00$\pm$0.042} & 83.90$\pm$0.016&82.00$\pm$0.048 &    \textbf{84.00$\pm$0.007}  \\
& R=1.0 & 53.00$\pm$0.021 & 78.50$\pm$0.093 & \underline{83.90$\pm$0.011}&\underline{83.90$\pm$0.011} &    \textbf{84.00$\pm$0.008}  \\
\bottomrule
\end{tabular}
\end{table}

Under low-noise conditions (R$=0.1, 0.2, 0.3$), LSQMM performs comparably to SMM and MLRSTM; under medium-to-high noise conditions (R$\in [0.4:0.1:0.7]$), LSQMM achieves optimal performance at four noise levels; under extreme noise conditions (R$=0.8, 0.9, 1.0$), LSQMM maintains approximately 84\% accuracy, significantly outperforming other methods. Comparative analysis reveals that LIBSVM is most sensitive to noise, with accuracy dropping to 53.00\% at R$=1.0$; SMM performs well under low noise but decays rapidly; TT-MMK shows good robustness but incurs the highest computational cost; MLRSTM remains stable under moderate noise but is slightly inferior to LSQMM under high noise conditions.

The superior performance of LSQMM stems from its quaternion matrix representation that preserves inter-channel correlations and the nuclear norm regularization that suppresses noise effects, providing an optimal balance between performance and robustness for noise-sensitive application scenarios.

\subsection{Parameter Sensitivity }
In this subsection, we analyze the influence of parameter variations on the classification performance of our model. The soft margin parameter \( C \) controls the penalty strength for misclassified samples in the loss function, while the parameter \( \lambda \) imposes a low-rank constraint on $\mathbf{W}$ via nuclear norm regularization.

\begin{figure*}[htbp]
    \centering
    \begin{subfigure}[b]{0.48\textwidth}
        \centering
        \includegraphics[width=\textwidth]{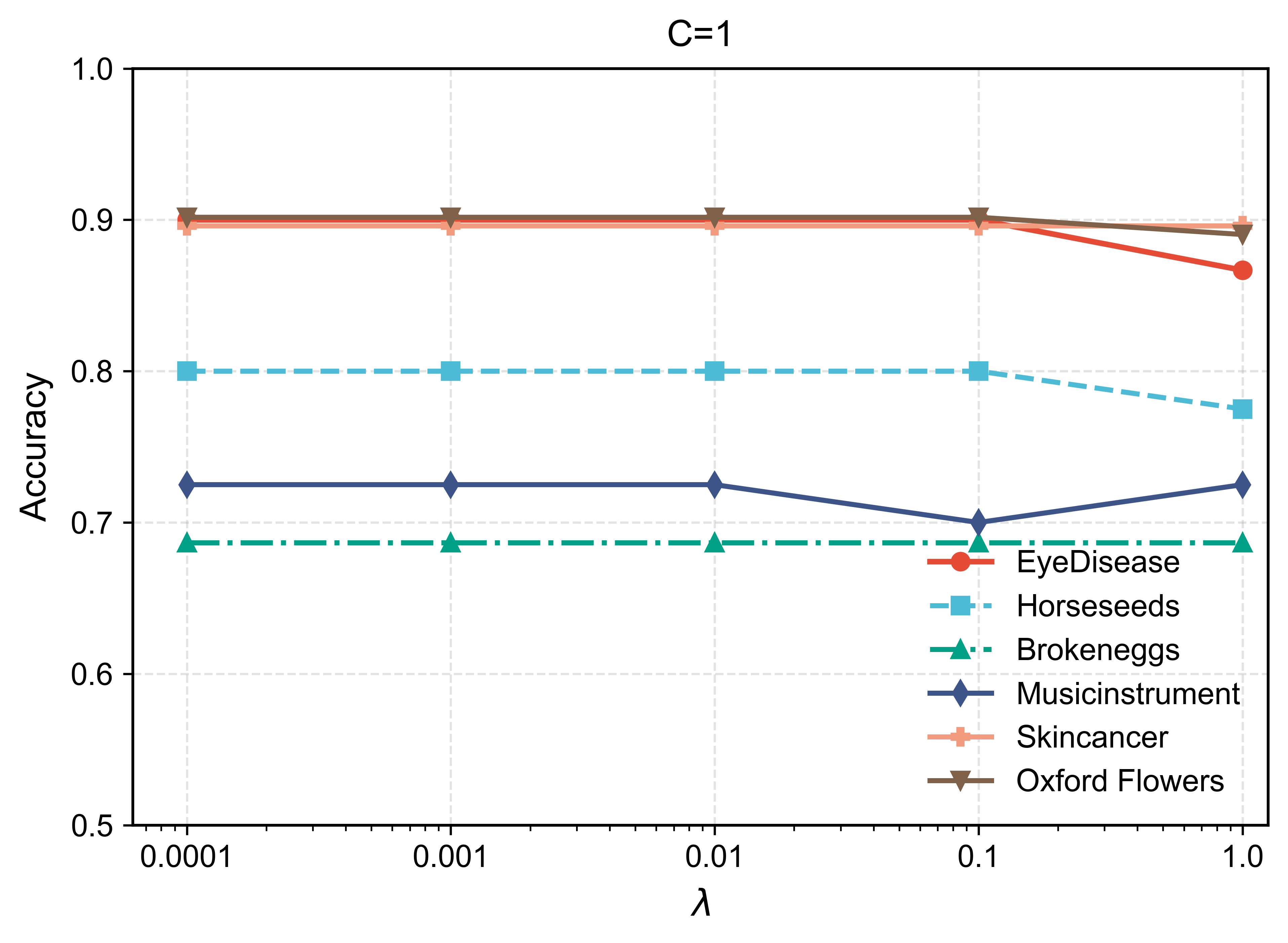}
        \caption{$C=1$}
        \label{fig:C1}
    \end{subfigure}
    \hfill
    \begin{subfigure}[b]{0.48\textwidth}
        \centering
        \includegraphics[width=\textwidth]{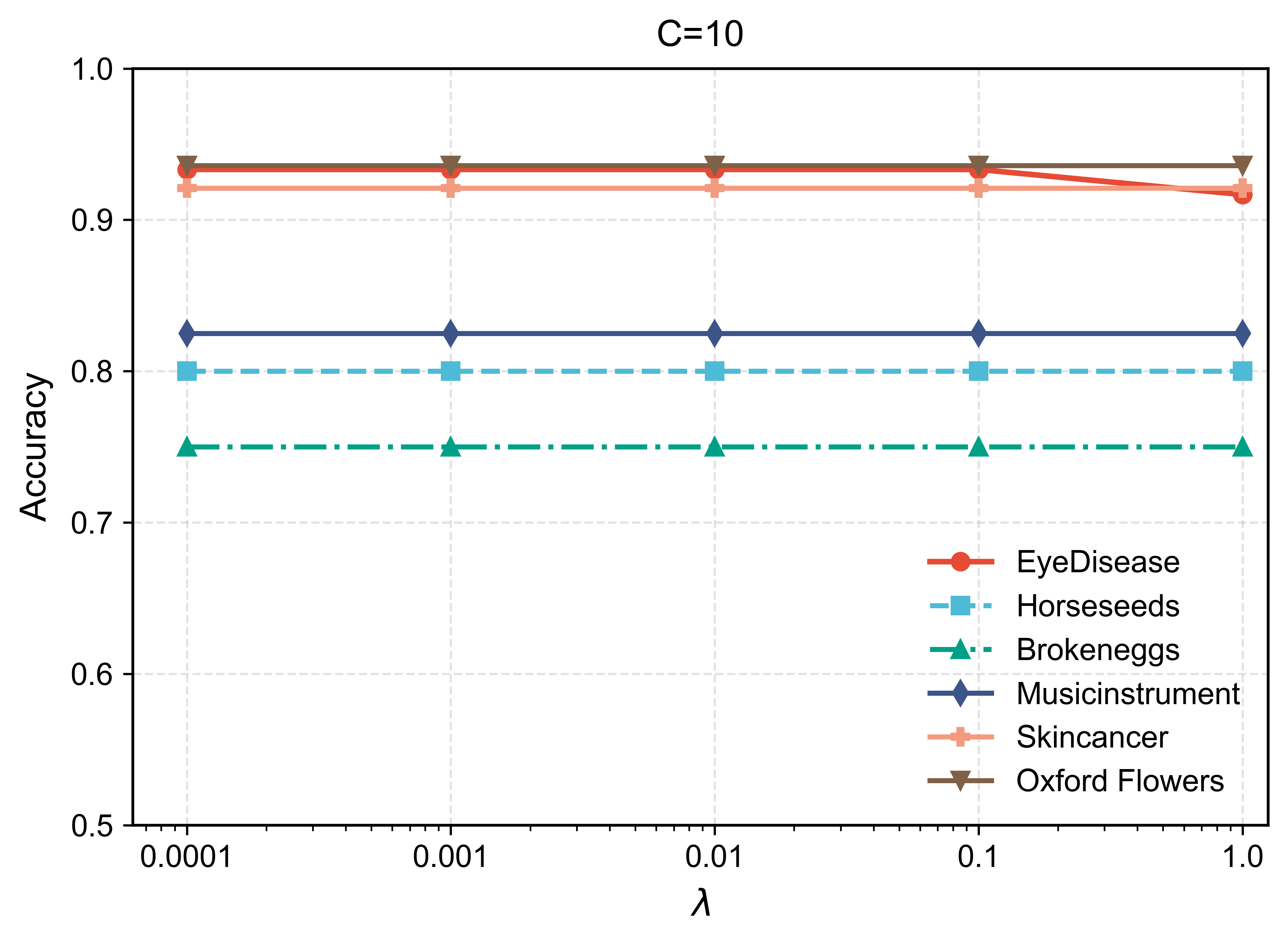}
        \caption{$C=10$}
        \label{fig:C10}
    \end{subfigure}

    \begin{subfigure}[b]{0.48\textwidth}
        \centering
        \includegraphics[width=\textwidth]{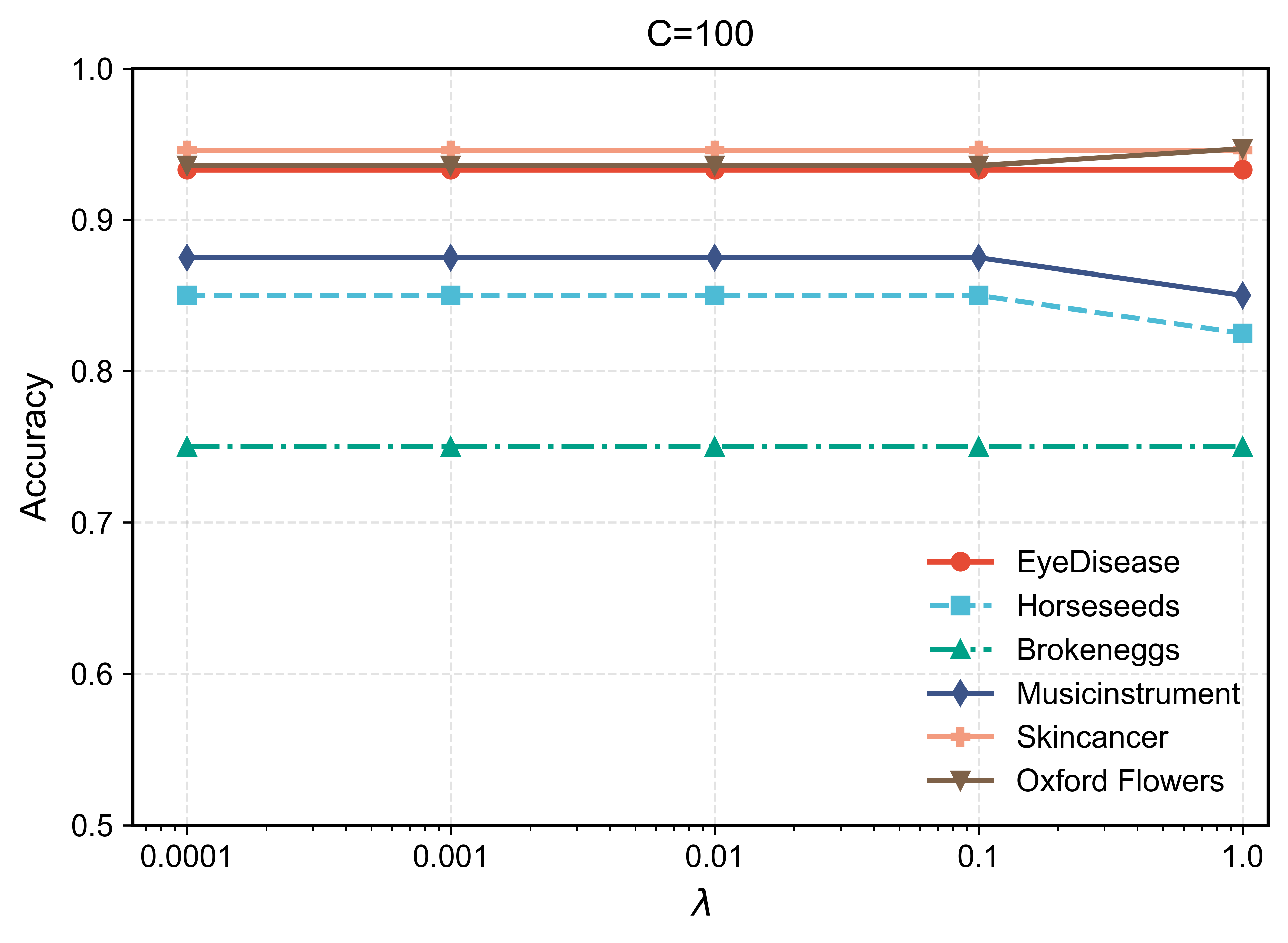}
        \caption{$C=100$}
        \label{fig:C100}
    \end{subfigure}
    \hfill
    \begin{subfigure}[b]{0.48\textwidth}
        \centering
        \includegraphics[width=\textwidth]{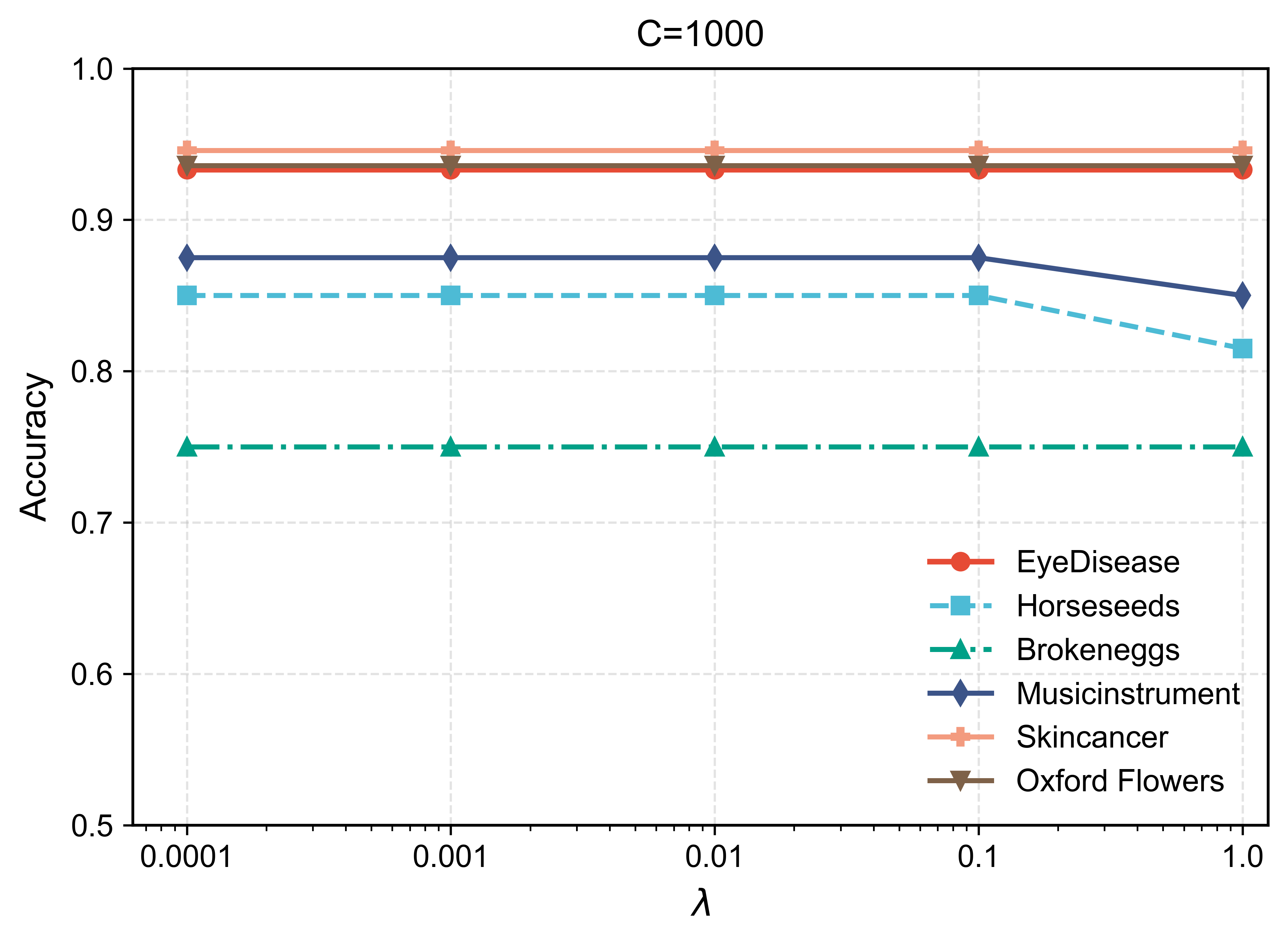}
        \caption{$C=1000$}
        \label{fig:C1000}
    \end{subfigure}

    \caption{
        Comparison of classification accuracy across different regularization parameter ($\lambda$) .
    }
    \label{fig:comparison}
\end{figure*}

As shown in Figure \ref{fig:comparison}, when $C$ is fixed and $ \lambda $ varies within a given range, all six color image datasets maintain stable and high classification accuracy. This demonstrates the excellent parameter robustness of the LSQMM model. Specifically, the model performance shows remarkable insensitivity to variations in the nuclear norm regularization parameter $ \lambda $, with all datasets exhibiting only minor fluctuations in accuracy.The introduction of nuclear norm regularization enables LSQMM to possess adaptive rank adjustment capability. The model can automatically learn appropriate low-rank feature representations according to the inherent structural characteristics of input data, thereby maintaining stable feature extraction performance under different regularization strengths.

Experimental results verify that LSQMM exhibits both good parameter robustness and adaptive rank selection capability in color image classification tasks. This not only simplifies parameter setting in practical applications but also validates the effectiveness of nuclear norm regularization in feature learning.

\begin{figure*}[htbp]
    \centering
    \begin{subfigure}[b]{0.48\textwidth}
        \centering
        \includegraphics[width=\textwidth]{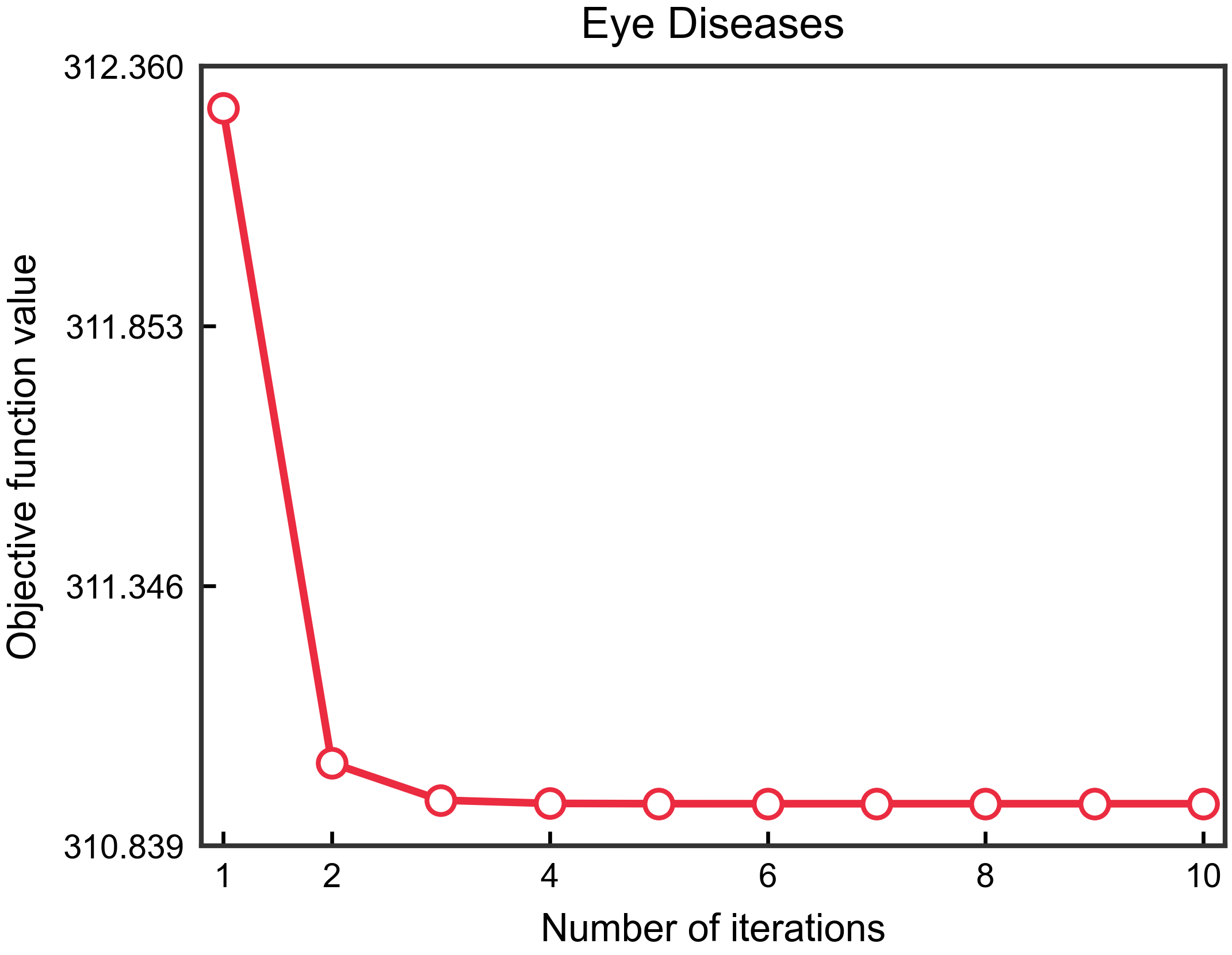}
        \caption{}
        \label{obj_eye}
    \end{subfigure}
    \hfill
    \begin{subfigure}[b]{0.48\textwidth}
        \centering
        \includegraphics[width=\textwidth]{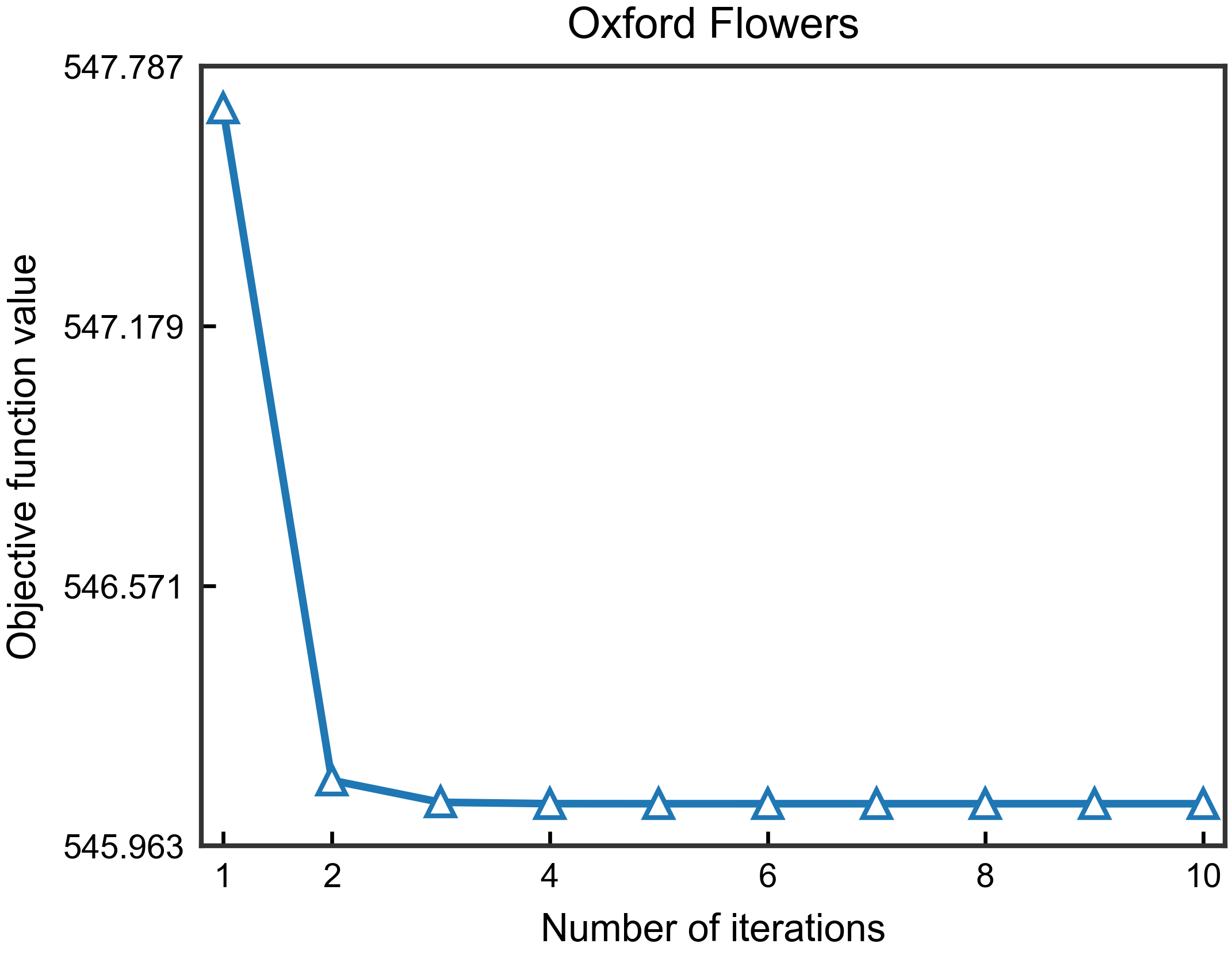}
        \caption{}
        \label{obj_flower}
    \end{subfigure}

    \caption{
        The values of objective functions versus the number of iterations on the  dataset.(a) the Eye Diseases dataset,(b)  the Oxford Flowers dataset .
    }
    \label{fig:comparison on datasets}
\end{figure*}

To experimentally validate the convergence behavior of Algorithm 1, Fig. \ref{fig:comparison on datasets} illustrates the evolution of the objective function values for LSQMM across two distinct datasets. As observed, the objective function values decrease rapidly with increasing iterations, demonstrating efficient convergence. Notably, LSQMM achieves a steady state within merely 10 iterations for all datasets, confirming its fast convergence and numerical stability. This consistent behavior underscores the effectiveness of the proposed optimization framework.

\section{Conclusion}
\label{sec:5}

To enhance the performance of color image classification, this paper proposes a novel LSQMM and its corresponding ADMM algorithmic framework. Unlike existing real-valued models, LSQMM directly models color images as quaternion matrices for model input, fully leveraging the inherent advantages of quaternion algebra in representing multi-channel color images. By treating the three RGB channels as an integrated entity, it effectively preserves the intrinsic coupling relationships and structural integrity among the channels. The introduction of the quaternion matrix nuclear norm as a low-rank regularization term enables LSQMM to thoroughly extract discriminative features from the images. Algorithmically, an iterative optimization scheme based on ADMM is designed, and the convergence of the algorithm is demonstrated both theoretically and experimentally. Experimental results on six benchmark color image datasets show that LSQMM outperforms existing support vector machine-type methods in terms of classification accuracy, F1-Score, and noise robustness, particularly excelling in small-sample high-dimensional scenarios.

Future research directions can be explored from the following aspects: First, sparsity regularization terms could be introduced to further enhance the model's feature selection capability and robustness; second, extending the model from the quaternion matrix framework to the quaternion tensor framework is worthwhile to handle higher-dimensional data processing tasks such as color videos; third, it deserves to design more efficient quaternion singular value decomposition algorithms to reduce computational complexity such as the stochastic or distributional schemes.

\bibliographystyle{plain}
\bibliography{main}

\end{document}